\documentclass{article}
    \PassOptionsToPackage{numbers, compress}{natbib}

    \usepackage[preprint]{neurips_2024}

\usepackage[utf8]{inputenc} 
\usepackage[T1]{fontenc}    
\usepackage{hyperref}       
\usepackage{url}            
\usepackage{booktabs}       
\usepackage{amsfonts}       
\usepackage{nicefrac}       
\usepackage{microtype}      
\usepackage{xcolor}         
\usepackage{graphicx}
\usepackage{float}
\usepackage[title]{appendix}
\usepackage{tcolorbox}
\usepackage{multirow}
\usepackage{amssymb}
\usepackage{bm}
\usepackage{pifont}
\usepackage{adjustbox}
\usepackage{subcaption}
\newcommand{\xmark}{\ding{55}}  
\newcommand{\newcheckmark}{\ding{51}}  

\title{LLM-Optic: Unveiling the Capabilities of Large Language Models for Universal Visual Grounding}

\author{Haoyu Zhao\textsuperscript{$1$}\quad
        Wenhang Ge\textsuperscript{$1$}\quad
        Ying-Cong Chen\textsuperscript{$1,2$}\thanks{Corresponding author} \quad \\
            \small$^1$ The Hong Kong University of Science and Technology~(Guangzhou).\quad \\
            \small$^2$ The Hong Kong University of Science and Technology.\\
            \small\texttt{haoyu-zhao@outlook.com, gewenhang01@gmail.com, yingcongchen@ust.hk} \\
}

\begin{document}
\maketitle

\begin{abstract}
\label{abstract}
Visual grounding is an essential tool that links user-provided text queries with query-specific regions within an image. Despite advancements in visual grounding models, their ability to comprehend complex queries remains limited. To overcome this limitation, we introduce LLM-Optic, an innovative method that utilizes Large Language Models (LLMs) as an optical lens to enhance existing visual grounding models in comprehending complex text queries involving intricate text structures, multiple objects, or object spatial relationships---situations that current models struggle with. LLM-Optic first employs an LLM as a Text Grounder to interpret complex text queries and accurately identify objects the user intends to locate. Then a pre-trained visual grounding model is used to generate candidate bounding boxes given the refined query by the Text Grounder. After that, LLM-Optic annotates the candidate bounding boxes with numerical marks to establish a connection between text and specific image regions, thereby linking two distinct modalities. Finally, it employs a Large Multimodal Model (LMM) as a Visual Grounder to select the marked candidate objects that best correspond to the original text query. Through LLM-Optic, we have achieved universal visual grounding, which allows for the detection of arbitrary objects specified by arbitrary human language input. Importantly, our method achieves this enhancement without requiring additional training or fine-tuning. Extensive experiments across various challenging benchmarks demonstrate that LLM-Optic achieves state-of-the-art zero-shot visual grounding capabilities. Project Page: {\color{blue}\href{https://haoyu-zhao.github.io/LLM-Optic.github.io/.}{https://haoyu-zhao.github.io/LLM-Optic.github.io/.}}
\end{abstract}

\section{Introduction} \label{intro}
\begin{figure}[t]
\begin{center}
\includegraphics[width=1\textwidth]{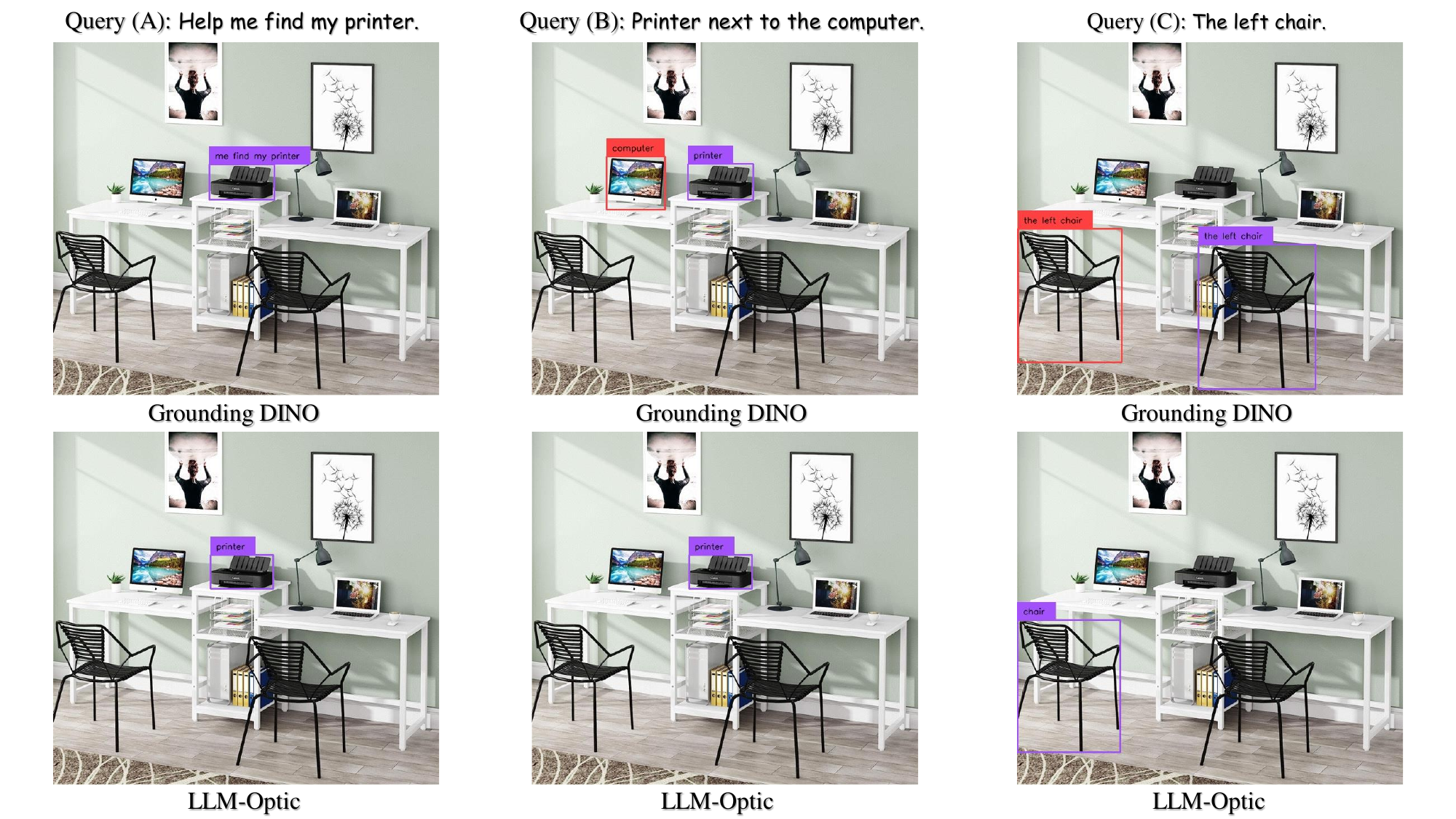}
\caption{LLM-Optic enhances the capabilities of the leading visual grounding model, Grounding DINO, by integrating the reasoning abilities of Large Language Models (LLMs), thereby achieving superior accuracy in visual grounding within any given query. Specifically, Grounding DINO exhibits limitations in the following areas: (1) it struggles with complex sentence structures, as demonstrated in Query (A); (2) it faces challenges with queries involving multiple objects and often fails to distinguish the primary object from its landmarks for precise localization (Query (B)); (3) it incorrectly interprets spatial relationships (Query (C)). However, our framework effectively addresses these issues.}
\label{fig:intro}
\end{center}
\end{figure}

Visual grounding is a pivotal task in computer vision that serves as the foundation for multiple fields such as autonomous driving \cite{autonomousdriving1,autonomousdriving2}, robotics \cite{robot1}, unmanned aerial vehicle (UAV) navigation \cite{uav1,uav2}. Pioneering studies \cite{groundingdino,yao2022detclip,zang2022open,zhong2022regionclip,kamath2021mdetr,li2022grounded,zhang2022glipv2} have conducted extensive explorations, continuously driving advancements in this task. Notably, Grounding DINO \cite{groundingdino} has exhibited superior performance, achieving state-of-the-art results. Grounding DINO is a formidable model for open-vocabulary object detection, distinguished by its superior performance and its capacity to process free-form text input queries. This capability arises from its implementation of vision-language modality fusion across multiple phases. However, despite the impressive achievements of Grounding DINO, it faces challenges in fully comprehending complex input text queries.  Its limitations become evident in various scenarios: (1) It struggles with complex sentence structures and misinterprets semantic information, as illustrated in Fig. \ref{fig:intro}(A), where it erroneously categorizes "me find my printer" as an object label; (2) It has difficulties with queries involving multiple objects, often failing to distinguish the primary object from its landmarks for precise localization, as shown in Fig. \ref{fig:intro}(B), where it incorrectly detects all mentioned objects in the query but neglects the main object of interest; (3) It incorrectly interprets spatial relationships, as demonstrated in Fig. \ref{fig:intro}(C). These limitations underscore the nuanced complexities of visual grounding and highlight the necessity for further enhancements. The primary reason that Grounding DINO exhibits these limitations is likely due to its use of BERT \cite{devlin2018bert} as its text encoder. BERT is primarily pre-trained through two tasks: Masked Language Modeling (MLM) and Next Sentence Prediction (NSP). Although these tasks facilitate the learning of the basic language structures, they are insufficient for capturing more complex linguistic phenomena and the intricacies of contextual relationships. In contrast, Large Language Models (LLMs) demonstrate superior capabilities in natural language understanding. LLMs are often pre-trained on a broader array of tasks and significantly larger datasets, which encompass complex text generation and comprehension. This extensive training enables a deeper understanding of complex semantic relationships and contextual variations thereby enhancing their capacity to interpret and respond to complex queries. Large Multimodal Models (LMMs), also known as MultiModal Large Language Models (MM-LLMs), represent an extension of LLMs, incorporating both image and text modalities. They surpass models like Grounding DINO in terms of language comprehension capabilities. However, LMMs currently exhibit limitations in visual grounding, often failing to generate precise bounding boxes for complex grounding queries. To enhance the performance of LMMs on visual language tasks such as visual grounding, current methods \cite{wang2023cogvlm, zhang2023llavagrounding,chen2023minigpt} involve training or fine-tuning existing large models. This process requires substantial training data, significant computational resource expenditure, and considerable training time.

In this work, we introduce LLM-Optic, a simple yet effective solution that enhances existing open-vocabulary object detection models by integrating the reasoning capabilities of Large Language Models, without additional training or fine-tuning. It effectively addresses the challenges that existing models encounter when interpreting complex text queries. Furthermore, LLM-Optic transcends the constraints typically associated with specialized models, which are often limited to narrowly defined tasks and specific output formats. It achieves \textbf{Universal Visual Grounding}, capable of identifying any number of objects based on varied descriptions and can also address situations where the described objects do not exist in the image, significantly enhancing the scope of visual grounding across diverse scenarios and achieving a level of visual grounding that aligns with human-like robustness. This framework achieves state-of-the-art accuracy in visual grounding benchmarks within a zero-shot setting, demonstrating improvements across all evaluated datasets with the highest increase observed at 22\% in the RefCOCOg \cite{refcocog} validation set. 

LLM-Optic primarily consists of three modules: an LLM-based Text Grounder, a Candidate Positioning and Setting Marks module, and an LMM-based Visual Grounder. Initially, an LLM functions as the Text Grounder, processing complex text queries to ascertain the true intent behind the text. The output from the Text Grounder is then fed into the Candidate Positioning and Setting Marks module. Within this module, the output of the Text Grounder, which is a simple yet precise description of the target, is relayed to a pre-trained open-vocabulary object detection model. This model is responsible for generating bounding boxes for candidate objects that could potentially correspond to the description. Each bounding box is then distinctly marked with a numerical identifier. Subsequently, the image with marked bounding boxes, together with the original query text, is processed by the Visual Grounder. Here, an LMM determines which of the marked objects correspond accurately to the query text description. Our framework's modular architecture enables us to seamlessly incorporate cutting-edge advancements and maintain a leading position in this domain.

In summary, our contribution can be summarized as:
\begin{itemize}
\item We propose LLM-Optic, a simple yet highly effective and fully modularized framework designed to enhance the capabilities of visual grounding models through three fundamental components: an LLM-based Text Grounder, which analyzes the underlying intent within the text query; a Candidate Positioning and Setting Marks module, responsible for producing potential target bounding boxes and establishing connections between text and corresponding image regions through marks; and an LMM-based Visual Grounder, which precisely identifies objects described in the query within the image. The principles of our framework are adaptable, enabling its application across a wide spectrum of computer vision tasks through a streamlined migration process
\item Our approach significantly extends the capabilities of current state-of-the-art grounding models without requiring additional training or fine-tuning. It effectively addresses the challenges that existing models encounter when interpreting complex text queries. Furthermore, our framework transcends the limitations of specialized models, which are often confined to narrowly defined tasks and restricted output formats, allowing for diverse input texts as well as varied output results. 
\item Extensive experiments across multiple visual grounding benchmarks show that the proposed framework significantly outperforms the state-of-the-art models in a zero-shot setting, without using additional data. These experiments show marked improvements on all evaluated datasets, with the most notable increase being 22\% in the RefCOCOg validation set.
\end{itemize}
\section{Related Work} \label{sec:related_work}
\paragraph{Visual Grounding.}
Visual grounding builds on object detection by linking specific regions of an image to natural language descriptions, enhancing understanding across modalities. The task of object detection, a core endeavor within the field of computer vision, continues to evolve. The primary objective of object detection is to achieve localization of all target objects that belong to predefined categories within an image \cite{dino,Deformable,fasterrcnn,Conditionaldetr,Dn-detr}. However, these Closed-Vocabulary object Detection (CVD) models, which focus on detection within a closed set of categories, struggle to generalize to novel classes due to the limitations imposed by these predefined categories. As research progresses, several related tasks have emerged. In addition to CVD, the field now distinguishes three tasks based on the type of input text: Open-Vocabulary object Detection (OVD), Referring Expression Comprehension (REC), and Phrase Grounding (PG). The OVD \cite{groundingdino,yao2022detclip,zang2022open,zhong2022regionclip,kamath2021mdetr,li2022grounded,zhang2022glipv2} task addresses the limitations of traditional object detection by enabling the identification of arbitrary classes beyond predefined categories. In contrast,  the REC \cite{yan2023universal,liu2023polyformer,luo2020multi,yang2019fast,kamath2021mdetr,recdeng2021transvg} task focuses on highly specific queries that may detail the relative positions, appearance characteristics, and other description of objects. Unlike OVD, where a single category label can correspond to multiple bounding boxes, in REC, each query specifically targets a unique object for localization. The PG \cite{sadhu2019zero,zhang2022glipv2,li2022grounded, dou2022coarse} task, meanwhile, requires locating all referenced objects (phrases) within a given sentence, necessitating comprehensive identification of multiple objects from the text input.

In our paper, we introduce LLM-Optic, which targets a more broadly applicable visual grounding task, namely Universal Visual Grounding. Unlike specialized models designed for specific tasks, such as those for the OVD task that struggles with complex user queries, or the REC task that can only locate a single object and requires specific training data, LLM-Optic is designed to be flexible. This means that LLM-Optic can identify any number of objects based on any given query, significantly expanding the scope and effectiveness of visual grounding across diverse scenarios, thus truly realizing universal visual grounding.

\paragraph{Large Multimodal Models.} Recent advances in the domain of Large Language Models (LLMs) have demonstrated remarkable outcomes. These achievements have rapidly extended to Large Multimodal Models (LMMs), which incorporate both text and image modalities. State-of-the-art models \cite{gpt3,gpt4v,team2023gemini,touvron2023llama,zhu2023minigpt,liu2023llava,liu2024llavanext, li2022blip, li2023blip} are extensively utilized across a diverse range of domains such as narrative generation \cite{zhuang2024vlogger,lin2023mm}, scene generation \cite{yu2023wonderjourney}, image captioning \cite{luo2024scalable}, and as evaluator for vision-language tasks \cite{wu2024gpt,zhang2023gpt}, demonstrating their remarkable capabilities. Moreover, LLMs and LMMs exhibit flexibility in various application methods. For example, some LMMs have been specifically trained or fine-tuned for tasks such as visual grounding \cite{wang2023cogvlm,zhang2023llavagrounding} and 3D understanding \cite{hong20233d,wang2023chat}.  Additionally, some studies \cite{Llm-grounder, wu2023autogen, schick2024toolformer, shen2024hugginggpt} have employed LLMs purely as agents for planning and tool-using through multi-turn conversations, without engaging in any additional training or fine-tuning. These diverse application methods showcase their significant practical value. 

However, state-of-the-art LMMs such as GPT-4V \cite{gpt4v} and LLaVA \cite{liu2023llava}, while being generalists across multiple domains, are not specifically optimized for visual grounding tasks. Consequently, they often struggle with complex visual grounding queries, typically failing to directly output accurate bounding boxes for the objects specified in the queries. Correspondingly, our framework, LLM-Optic, to the best of our knowledge, represents the first attempt to integrate the capabilities of LLMs and LMMs with 2D visual grounding models. This integration leverages the reasoning capabilities of LLMs and LMMs for both images and text, along with the precise localization abilities of visual grounding models. It operates without the need for any additional training or fine-tuning. Furthermore, it requires only a single-turn conversation per interaction, utilizing minimal token consumption.

\section{Method} \label{sec:method}
Our goal is to improve the capabilities of existing visual grounding models to understand complex text queries by integrating the advanced reasoning abilities of LLMs and LMMs. This enhancement enables them to overcome current limitations and thereby ensures accurate visual grounding in response to any query. Our proposed framework, named LLM-Optic, is structured into three main components: Text Grounder §\ref{Text Grounder}, Candidate Positioning and Setting Marks §\ref{Candidate Positioning and Setting Marks}, and Visual Grounder §\ref{Visual Grounder}. This highly modular approach requires no additional training or fine-tuning, and each component is interchangeable with any state-of-the-art model.
\begin{figure*}[t]
	\centering
			{\includegraphics[width=1\linewidth]{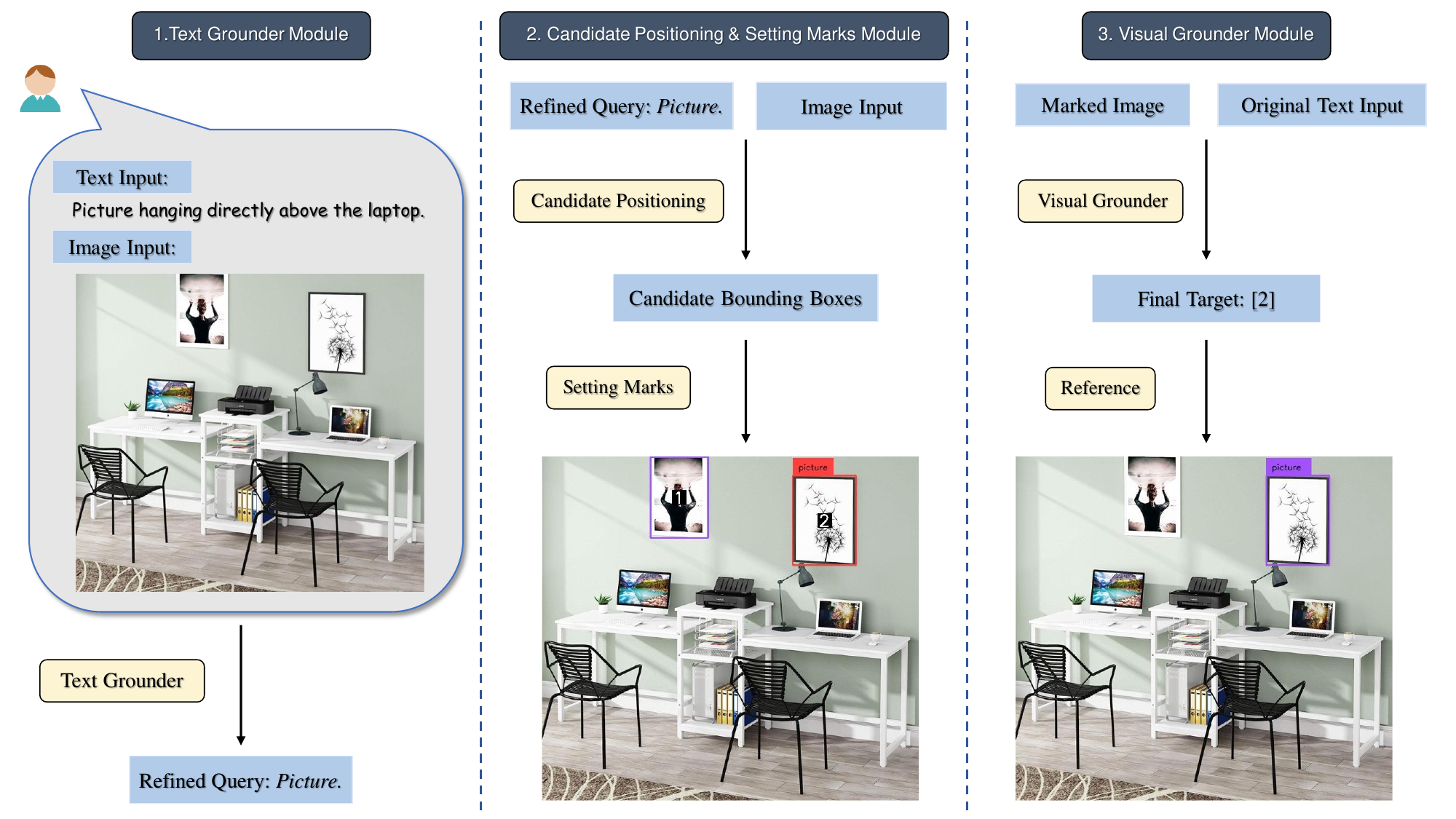}} 
	\caption{\textbf{Overview of LLM-Optic.} We propose using LLMs and LMMs as effective reasoning modules for handling complex user queries to achieve universal visual grounding. Our framework includes three key modules: an LLM-based Text Grounder, a Candidate Positioning and Setting Marks module, and an LMM-based Visual Grounder. It does not require any additional training and features a fully modular design, allowing for the seamless integration of rapid advancements in new technologies.}
	\label{fig:method}
\end{figure*}
\subsection{Text Grounder} \label{Text Grounder}
LLMs have demonstrated promising performance in natural language understanding \cite{gpt3,gpt32}. For complex user queries, we utilize an LLM (GPT-3.5 Turbo) as a Text Grounder to parse and interpret the text query. We inform the LLM of the expected input and output formats; details of the prompt used for the LLM are provided in Appendix \ref{appendix.A}. The Text Grounder effectively extracts key information, ensuring a deep understanding of the user's actual intent. This accurate interpretation is crucial as it guides the subsequent pre-trained visual grounding model to precisely identify the object specified in user queries. For instance, inputting the query "Picture hanging directly above the laptop" directly into a visual grounding model might lead to misinterpretation. However, the Text Grounder employs semantic and commonsense reasoning to determine that the focus should be on locating the picture, with the laptop serving merely as a reference landmark. After the Text Grounder processes the user's original query, the refined input to the visual grounding model becomes the extracted text "Picture", identifying a specific object category that the visual grounding model can efficiently handle.
\subsection{Candidate Positioning and Setting Marks} \label{Candidate Positioning and Setting Marks}
\begin{figure}[htbp]
    \centering
    \begin{minipage}{.31\textwidth}
        \centering
        \includegraphics[width=\linewidth]{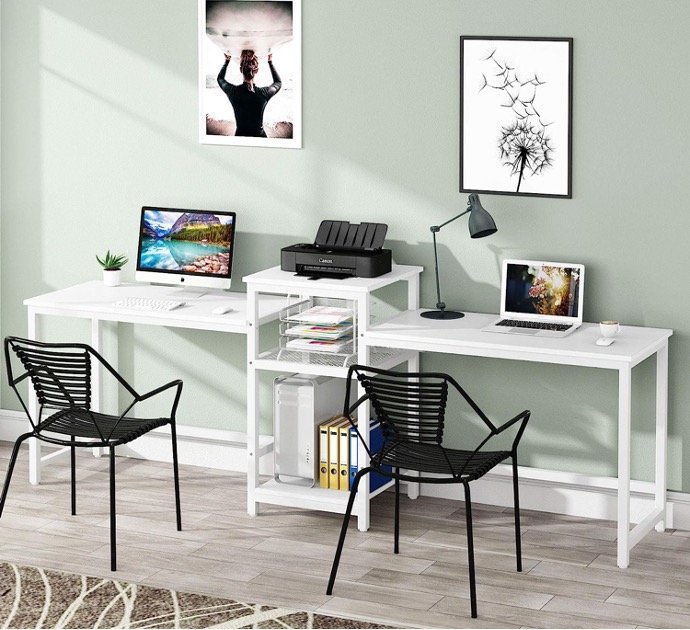}
        \captionsetup{skip=2pt}
        \caption*{(a)}
    \end{minipage}%
    \hfill
    \bm{$\rightarrow$}
    \hfill
    \begin{minipage}{.31\textwidth}
        \centering
        \includegraphics[width=\linewidth]{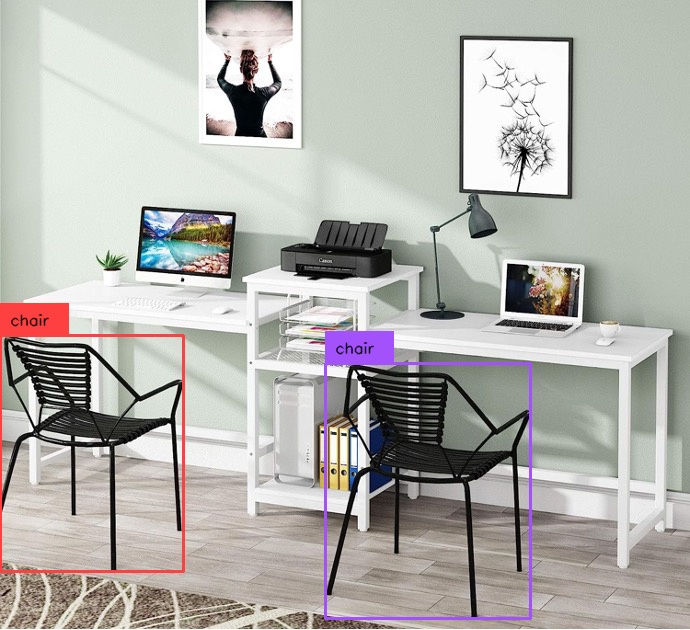}
        \captionsetup{skip=2pt}
        \caption*{(b)}
    \end{minipage}%
    \hfill
    \bm{$\rightarrow$} 
    \hfill
    \begin{minipage}{.31\textwidth}
        \centering
        \includegraphics[width=\linewidth]{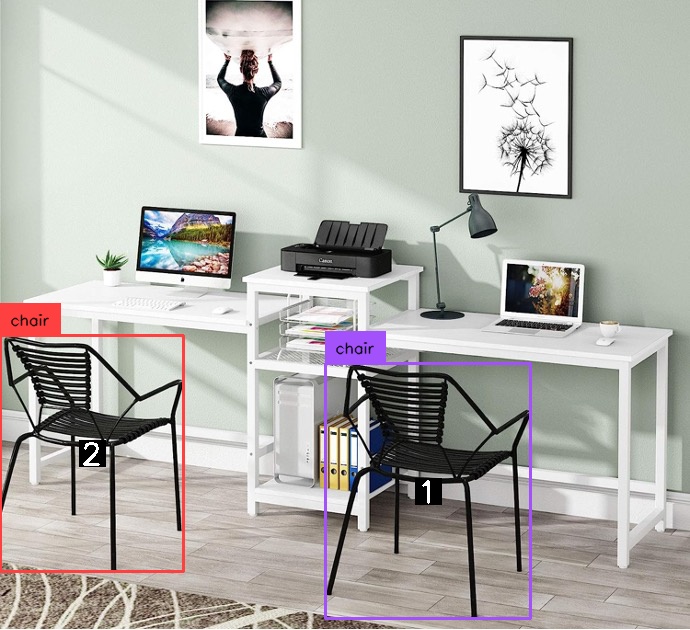}
        \captionsetup{skip=2pt}
        \caption*{(c)} 
    \end{minipage}
    \caption{\textbf{The Process of Candidate Positioning and Setting Marks.} Image (a) is the original input image, while Image (b) shows the image annotated with bounding boxes, and Image (c) displays the image after marks have been applied to each bounding box.}
    \label{fig:mark}
\end{figure}

\paragraph{Candidate Positioning.}
After processing with the Text Grounder, we derived a simple yet precise expression of the input query, termed "Refined Query", such as "Picture". This concise expression is then relayed to an open-vocabulary object detection model, specifically, we employ Grounding DINO. Grounding DINO is a robust, pre-trained, open-vocabulary object detector that exhibits superior performance across existing visual grounding models. This model is responsible for generating bounding boxes for candidate objects that could potentially match the refined query. However, these candidates are not the final targets; rather, they are preliminary selections that match the specified categories of the query object. Further processing, including the analysis of object appearance and spatial relationships, requires additional reasoning by the Visual Grounder in the subsequent module.
\paragraph{Setting Marks.}
After positioning candidates, we mark the center of each candidate bounding box with a unique numerical identifier, as illustrated in Fig. \ref{fig:mark}. These identifiers serve as unique identities for each of the candidate bounding boxes. This step has two primary purposes: First, it establishes a textual-visual link by directly indexing each specific region annotated with a bounding box to a corresponding number, enabling the subsequent Visual Grounder to more effectively reason and respond based on these marks. Second, according to the study from Set-of-Mark Visual Prompting \cite{setofmark}, these marks can effectively reduce the hallucination commonly encountered in Large Multimodal Models, thereby enhancing the accuracy of the Visual Grounder.

\subsection{Visual Grounder} \label{Visual Grounder}
\begin{figure}[ht] 
\centering
\begin{tcolorbox}
\begin{minipage}{0.45\textwidth} 
  \centering 
  \includegraphics[width=0.9\linewidth]{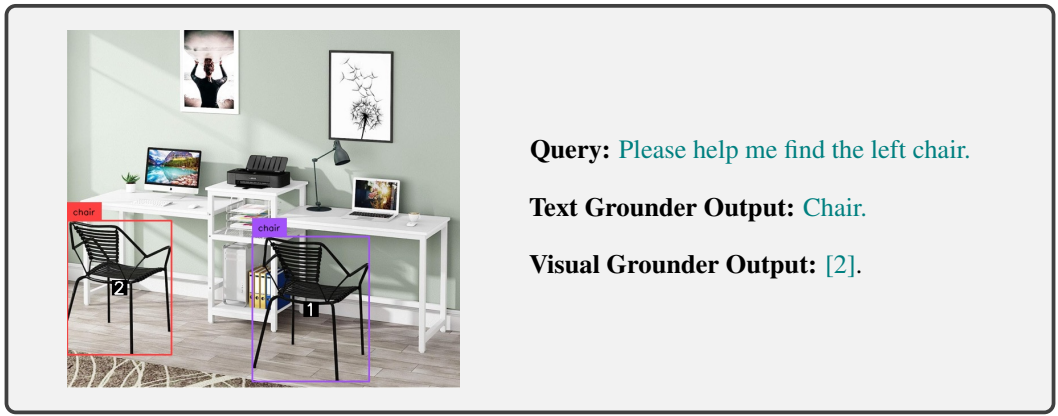}
\end{minipage}%
\hfill 
\begin{minipage}{0.5\textwidth} 
    \textbf{Query:} \textcolor{teal}{Please help me find the left chair.}\\\\
    \textbf{Text Grounder Output:} \textcolor{teal}{Chair.}\\\\
    \textbf{Visual Grounder Output:} \textcolor{teal}{[2]}.
\end{minipage}
\end{tcolorbox} 
\caption{\textbf{An example of Text Grounder and Visual Grounder output.} We have increased the size of the marks to enhance visibility; however, the actual marks are smaller, as shown in the Additional Results in Appendix \ref{appendix.E}, to avoid obscuring the target object.}
\label{fig:visualgrounder}
\end{figure}

After the image has been marked, it is paired with the original unprocessed query to form an image-text pair. This pair is then input into an LMM, specifically GPT-4V in LLM-Optic, which functions as the Visual Grounder. State-of-the-art LMMs, such as GPT-4V, have demonstrated remarkable effectiveness in general vision-language tasks \cite{yang2023dawn,earlyevaluation}. As the Visual Grounder, the LMM utilizes its reasoning capabilities to analyze the marked image and original text query, ultimately selecting objects that closely match the target described in the query among the marked candidates. Similarly to the Text Grounder, we inform the LMM of the expected input and output formats; details of the prompt used for the LMM are provided in Appendix \ref{appendix.A}. The output from the LMM is the marking identifier of the objects that match the text query, which is then used to index the previously saved bounding boxes to locate the target object. A detailed example is illustrated in Fig. \ref{fig:visualgrounder}.

\definecolor{darkgreen}{rgb}{0.0, 0.5, 0.0}

\section{Experiments} \label{sec:Experiments}
\subsection{Datasets}
To evaluate the performance of LLM-Optic, we conducted experiments across multiple datasets, including RefCOCO \cite{refcoco}, RefCOCOg \cite{refcocog}, and the Description Detection Dataset ($D^{3}$) \cite{d3}. These datasets, characterized by their complex descriptions, are widely utilized for the training and testing of existing visual grounding models.

\paragraph{RefCOCO \& RefCOCOg.}
RefCOCO \cite{refcoco} and RefCOCOg \cite{refcocog} are datasets specifically designed for the Referring Expression Comprehension (REC) task, which focuses on understanding natural language expressions that denote a unique object within an image. Originating from MS-COCO \cite{mscoco}, RefCOCO is divided into four splits: Train, TestA, TestB, and Val, while RefCOCOg is divided into three: Train, Test, and Val. A key difference between the two is the complexity of the referring expressions; RefCOCO accommodates any type of language, whereas RefCOCOg provides more detailed descriptions of objects. Notably, the average expression length in RefCOCOg is 8.4 words, significantly longer than the 3.5 words typical in RefCOCO.

\paragraph{Description Detection Dataset ($D^{3}$).}
Distinct from RefCOCO and RefCOCOg, the $D^{3}$ Dataset employs diverse and flexible language expressions that vary in length and complexity. A notable feature of $D^{3}$ is its incorporation of numerous descriptions that highlight the absence of certain concepts, such as "helicopter \textbf{not} flying in the air". This inclusion significantly aids in evaluating the robustness of different methods.

\subsection{Evaluation Metrics}
We evaluate the visual grounding capabilities of different models using multiple commonly used metrics, including the mean Intersection-over-Union (mIoU), as well as Accuracy@0.25 and Accuracy@0.5. Accuracy@0.25 and Accuracy@0.5 refer to the accuracies of bounding box predictions where the Intersection-over-Union (IoU) with the ground-truth bounding box exceeds the thresholds of 0.25 and 0.5, respectively.

\subsection{Baselines}
We select GPT-4V \cite{gpt4v}, Grounding DINO \cite{groundingdino}, and UNINEXT \cite{uninext} as our baselines due to their exemplary representativeness and robustness. In terms of visual grounding, Grounding DINO and UNINEXT have demonstrated state-of-the-art performance in their respective tasks, specifically in the OVD task for Grounding DINO and the REC task for UNINEXT. Likewise, GPT-4V also excels as the state-of-the-art LMM. Notably, GPT-4V and Grounding DINO have not been trained on the REC dataset (RefCOCO series). In contrast, UNINEXT has undergone training on the RefCOCO series, demonstrating exemplary state-of-the-art performance on it. We include this baseline to demonstrate the capabilities of a \textbf{trained} pipeline, establishing a performance ceiling for the RefCOCO series compared to our \textbf{zero-shot} setting.  Detailed specifications of the baselines are provided below.

\paragraph{GPT-4V.}
GPT-4V(ison) \cite{gpt4v} is a \textbf{state-of-the-art} Large Multimodal Model that currently exhibits exceptional performance across various tasks \cite{yang2023dawn,earlyevaluation}. In our experiments, we provided GPT-4V with a carefully designed prompt, which is detailed in Appendix \ref{appendix.A}, enabling it to directly output bounding boxes for target objects based on the input user query.

\paragraph{Grounding DINO.}
Grounding DINO \cite{groundingdino} is a robust visual grounding model that demonstrates \textbf{state-of-the-art} performance across various object detection datasets. The model enhances the closed-set detector, DINO \cite{dino}, by integrating a dual-encoder-single-decoder architecture, which facilitates vision-language modality fusion at multiple stages. This advanced architecture comprises a feature enhancer, a language-guided query selection module, and a cross-modality decoder.

\paragraph{UNINEXT.}
UNINEXT \cite{uninext} is a model that achieves \textbf{state-of-the-art} results on the RefCOCO series using additional training data. It is structured around three primary components: prompt generation, image-prompt feature fusion, and object discovery and retrieval. Its training process unfolds in three stages: general perception pre-training, image-level joint training, and video-level joint training. Notably, the RefCOCO series is employed for fine-tuning during the latter two stages.

\subsection{Results}
Due to quota restrictions of GPT-4V, we adopted a sampling approach for our experiments. We randomly sampled 200 text-image pairs from each split of RefCOCO (800 total), 200 text-image pairs from each split of RefCOCOg (600 total), and 200 text-image pairs from $D^{3}$. In total, these sampling strategies yielded 1,600 text-image pairs, providing a substantial and diverse testing dataset for our experiments. Furthermore, preliminary tests indicated GPT-4's relatively weaker grounding capabilities in handling complex queries; therefore we only sampled 50 text-image pairs from each dataset split for GPT-4V, totaling 400 text-image pairs. All experiments were conducted in a consistent environment with uniform settings.

We conducted experiments following the dataset settings described above, and the results are detailed in Table \ref{tab:performance_comparison}. These results clearly demonstrate that LLM-Optic achieves state-of-the-art performance in a zero-shot setting across all evaluated datasets, surpassing Grounding DINO significantly. For instance, LLM-Optic shows a notable improvement of 22\% on the RefCOCOg validation set in terms of Accuracy@0.5. Compared to the fine-tuned state-of-the-art model UNINEXT, which is typically trained with 32 or 16 A100 GPUs along with additional training data, LLM-Optic's performance is comparable, even though our framework does not require additional training. Additionally, in the $D^{3}$ dataset, where UNINEXT was not fine-tuned, LLM-Optic's performance exceeds that of UNINEXT by 20\%, underscoring our framework's effectiveness. In addition to its outstanding performance, it is also important to note that to ensure fair comparisons, we employed specific settings that potentially indicate that the actual performance of LLM-Optic might be higher than what is presented in Table \ref{tab:performance_comparison}. For more detailed information, please refer to Appendix \ref{appendix.B}.

\begin{table}[htbp]
\centering
\small
\caption{Comparison with state-of-the-art baselines on the RefCOCO, RefCOCOg, and $D^{3}$ datasets highlights our superior performance, with the highest scores presented \textbf{in bold}. Results for UNINEXT on the RefCOCO series are included to demonstrate the capabilities of the current trained pipeline, serving as a ceiling on the RefCOCO series against our zero-shot setting. Our framework significantly outperforms UNINEXT under the same zero-shot settings on the $D^{3}$ dataset, illustrating our superior performance and robustness. The results clearly indicate that our framework surpasses these state-of-the-art methods by a large margin. * indicates whether the model was fine-tuned on the RefCOCO series.}

\label{tab:performance_comparison}
\resizebox{\textwidth}{!}{
\begin{tabular}{@{}ccccccccccc@{}} 
\toprule
\multirow{2}{*}{\textbf{Method}} & \multirow{2}{*}{\textbf{Fine-Tuning*}} & \multirow{2}{*}{\textbf{Metric} $\uparrow$} & \multicolumn{4}{c}{\textbf{RefCOCO}} & \multicolumn{3}{c}{\textbf{RefCOCOg}} & \multirow{2}{*}{\textbf{$D^{3}$}}\\
\cmidrule(lr){4-7} \cmidrule(lr){8-10} 
&  &  & Train & TestA & TestB & Val& Train & Test & Val \\ \midrule
\multirow{3}{*}{UNINEXT \cite{uninext}} & \multirow{3}{*}{\newcheckmark}
& mIoU & 0.917 & 0.853 & 0.812 & 0.852 & 0.901 & 0.798 & 0.792 & 0.502 \\
& & Acc@0.25 & 0.975& 0.920 & 0.890 & 0.935 & 0.955 & 0.865 & 0.885 & 0.550 \\
& & Acc@0.5 & 0.965 & 0.915 & 0.880 & 0.895 & 0.945 & 0.835 & 0.835 & 0.515 \\ \midrule \midrule
\multirow{3}{*}{GPT-4V \cite{gpt4v}} & \multirow{3}{*}{\xmark}
& mIoU & 0.227 & 0.131 & 0.207 & 0.160 & 0.177 & 0.136 & 0.167 & 0 \\
& & Acc@0.25 & 0.400 & 0.260 & 0.380  & 0.320 & 0.280 & 0.260  & 0.340 & 0 \\
& & Acc@0.5 & 0.140 & 0.020 & 0.060 & 0.040 & 0.120 & 0.020 & 0.060 & 0 \\  \midrule
\multirow{3}{*}{Grounding DINO \cite{groundingdino}} & \multirow{3}{*}{\xmark}
& mIoU & 0.482 & 0.543 & 0.353& 0.444 & 0.549 & 0.579 & 0.508 & 0.641\\
& & Acc@0.25 &0.550 & 0.635& 0.430 & 0.530 & 0.615 & 0.665 & 0.610 & 0.715 \\
& & Acc@0.5 & 0.480 & 0.560& 0.340 & 0.435 & 0.560 & 0.585 & 0.505 & 0.680 \\ 
\midrule
\multirow{3}{*}{LLM-Optic} & \multirow{3}{*}{\xmark}
& mIoU & \textbf{0.576} &\textbf{0.577} & \textbf{0.520} & \textbf{0.500} & \textbf{0.617} & \textbf{0.683} & \textbf{0.620} & \textbf{0.681}\\
& & Acc@0.25 & \textbf{0.675} & \textbf{0.650} &\textbf{0.605} & \textbf{0.560} & \textbf{0.685} & \textbf{0.760} & \textbf{0.645} & \textbf{0.730} \\
& & Acc@0.5 & \textbf{0.590} &\textbf{0.585} & \textbf{0.535} & \textbf{0.500} & \textbf{0.635} & \textbf{0.705} & \textbf{0.725} & \textbf{0.710} \\ 
\bottomrule
\end{tabular}
}
\end{table}

\subsection{Additional Evaluation}
\paragraph{Ablation Study of using different LLMs \& LMMs.}
We conducted an ablation study on the $D^{3}$ dataset to evaluate the effectiveness of different LLMs as Text Grounders and LMMs as Visual Grounders. We randomly selected 100 samples from the $D^{3}$ dataset for this analysis. For the Text Grounder, we employed various LLMs, including GPT-3.5 Turbo, GPT-4, Llama-2 \cite{touvron2023llama}, and Llama-3 \cite{touvron2023llama}. Our results, presented in Table \ref{tab:differgpt}, indicate that all tested LLMs exhibit robust performance. Notably, the performance of open-sourced LLMs was comparable to that of GPT-4, with even the basic 7B model demonstrating sufficient capability to serve as a Text Grounder for input queries. 

For the Visual Grounder, in addition to GPT-4V, we also evaluated the widely-used open-sourced LMMs, LLaVa-1.5 \cite{liu2023llava} and LLaVa-1.6 (also known as LLaVa-Next) \cite{liu2024llavanext}, with different model parameter sizes. These evaluations indicated that while LLaVa-1.6 was capable of functioning as the Visual Grounder, it was less effective in providing accurate responses compared to GPT-4V. This is likely because GPT-4V has considerably more model parameters and was pre-trained on more extensive datasets. In contrast, LLaVa-1.5 struggled to function as required, failing to fulfill the task of a Visual Grounder. This underperformance is likely due to LLaVa-1.6 possessing enhanced visual reasoning and OCR capabilities, attributed to an improved visual instruction tuning data mixture, which LLaVa-1.5 lacks. This solidifies GPT-4V as the current most reliable option. Our modular design allows us to readily substitute components with the latest pre-trained models, thereby potentially enhancing our model's performance as research in the field progresses.

\begin{figure}[ht] 
\centering
\begin{tcolorbox}[boxrule=0mm, boxsep=0mm, colframe=gray]
\begin{minipage}[t]{.47\textwidth}
    \centering
    \fbox{\includegraphics[height=3.5cm,keepaspectratio]{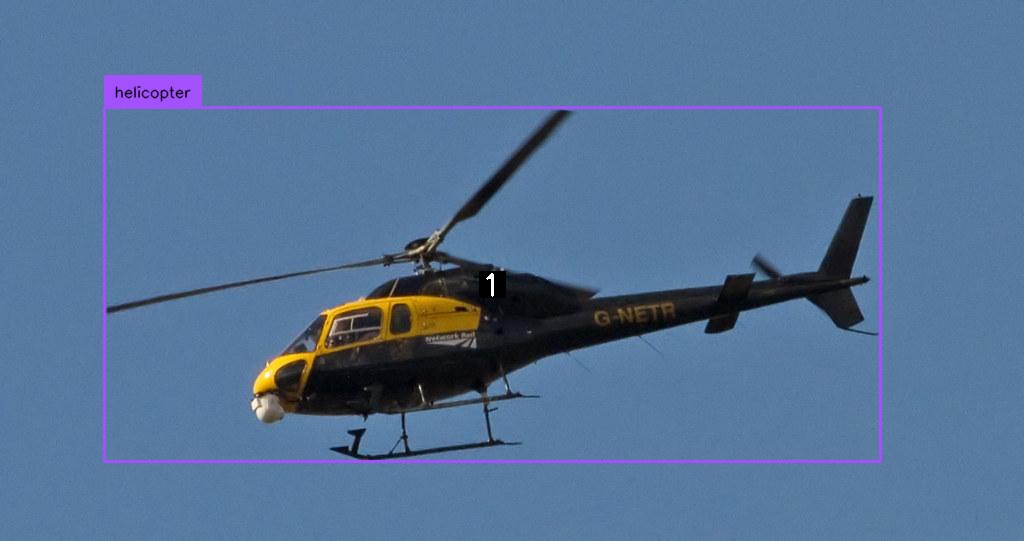}}\vspace{3mm}\\
    \small
    \raggedright 
    \textbf{\quad Situation}: \textcolor{teal}{Zero-Object}\\
    \textbf{\quad Query}: \textcolor{teal}{helicopter not flying in the air.}\\
     \mbox{\quad \textbf{Output}: \textcolor{teal}{There are no targets that fit the description.}}
\end{minipage}%
\hfill
\begin{minipage}[t]{.47\textwidth}
    \centering
    \fbox{\includegraphics[height=3.5cm,keepaspectratio]{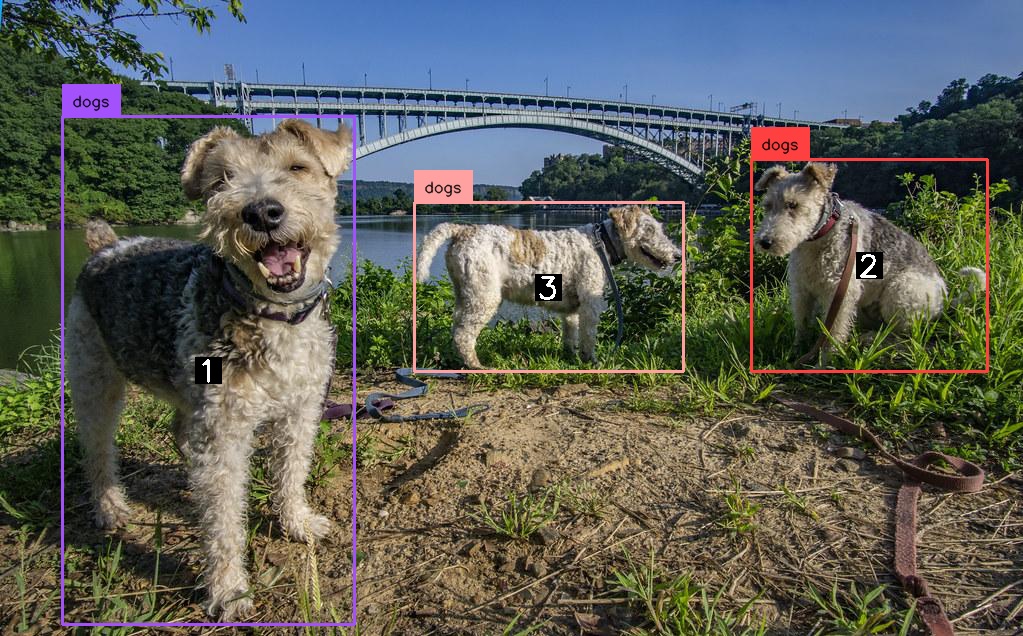}}\vspace{3mm}\\
    \raggedright 
    \small
    \textbf{\quad\quad Situation}: \textcolor{teal}{Multi-Object}\\
    \textbf{\quad\quad Query}: \textcolor{teal}{dogs with curly hair.}\\
    \textbf{\quad \quad Output}: \textcolor{teal}{{"Subject": [1,2,3]}}
\end{minipage}%
\end{tcolorbox}
\caption{LLM-Optic is capable of handling scenarios where the query does not correspond to any object in the image or corresponds to multiple objects within the image.}
\label{fig:more}
\end{figure}

\begin{table}[hb]
\centering
\small
\caption{An ablation study on the use of different Large Language Models (LLMs) as Text Grounders and Large Multimodal Models (LMMs) as Visual Grounders on the $D^3$ Dataset. * indicates whether the model was made open-source.}
\label{tab:differgpt}
\begin{tabular}{@{}ccccc@{}}
\toprule
\textbf{Text Grounder} & \textbf{Visual Grounder} & \textbf{mIoU$\uparrow$} & \textbf{Acc@0.25$\uparrow$} & \textbf{Acc@0.5$\uparrow$} \\
\midrule
GPT-3.5 Turbo & GPT-4V & 0.684 & 0.740 & 0.710 \\
GPT-4   &  GPT-4V   & 0.696 & 0.750 & 0.720 \\
Llama-2-7B* \cite{touvron2023llama}  &  GPT-4V    & 0.666 & 0.700 & 0.690 \\
Llama-3-7B* \cite{touvron2023llama}  &    GPT-4V     & 0.688 & 0.730 & 0.720 \\\midrule
Llama-2-7B* \cite{touvron2023llama}  &    LLaVA-1.6-34B* \cite{liu2024llavanext}    & 0.507 & 0.540 & 0.520 \\
Llama-3-7B* \cite{touvron2023llama}  &    LLaVA-1.6-34B* \cite{liu2024llavanext}       & 0.534 & 0.560 & 0.550 \\
Llama-3-7B* \cite{touvron2023llama}  &    LLaVA-1.6-Vicuna-13B* \cite{liu2023llava}       & 0.581 & 0.610 & 0.600 \\
\bottomrule
\end{tabular}
\end{table}
\paragraph{Robustness of LLM-Optic.}
LLM-Optic overcomes the typical limitations of specialized models, which are often restricted to narrowly defined tasks and specific output formats. For example, OVD models are limited by a lack of detailed contextual understanding, focusing primarily on predefined categories rather than considering specific attributes of targets in detail. However, REC models excel at parsing extended descriptions but are limited by the assumption that only one target is present in the image and also require specific training data. This assumption introduces limitations in scenarios with absent targets or multiple targets. Distinct from these models, LLM-Optic offers a versatile and robust solution capable of addressing a wide array of complex visual grounding challenges. LLM-Optic achieves \textbf{Universal Visual Grounding}, capable of identifying any number of objects based on varied descriptions, and can also address situations where the described objects do not exist in the image. This significantly enhances the scope of visual grounding across diverse scenarios, achieving a level of visual grounding comparable to human-like robustness. The distinctions among different visual grounding models are summarized in Table \ref{tab:methods_comparison}. Additionally, we present cases of LLM-Optic handling variable situations in Fig. \ref{fig:more}.

\begin{table}[tbp]
\centering 
\small
\caption{Comparisons between our proposed LLM-Optic and existing task-specific models highlight the broader applicability and superior robustness of LLM-Optic. Unlike models specifically designed for tasks such as Open-Vocabulary object Detection (OVD) and Referring Expression Comprehension (REC), LLM-Optic effectively handles variable situations and achieves Universal Visual Grounding.} 
\label{tab:methods_comparison}
\begin{tabular}{lccc} 
\toprule 
\textbf{Query Type} & \textbf{REC Models} & \textbf{OVD Models} & \textbf{LLM-Optic} \\
\midrule
Complex & \newcheckmark & \xmark & \newcheckmark \\
Multi-Object &  \xmark &\newcheckmark & \newcheckmark \\
Zero-Object &  \xmark &\newcheckmark & \newcheckmark \\
\bottomrule 
\end{tabular}
\end{table}

\definecolor{darkgreen}{rgb}{0.0, 0.5, 0.0}

\section{Conclusion} \label{conclusion}
In this paper, we introduce LLM-Optic, a novel framework designed to enhance the capabilities of current state-of-the-art visual grounding models without requiring additional training or fine-tuning. It effectively addresses the limitations that current models encounter when interpreting complex text queries. Furthermore, LLM-Optic overcomes the limitations inherent in specialized models, which are typically constrained by narrowly defined tasks and output formats, achieving Universal Visual Grounding. The framework features a modular design, with each component being interchangeable with any state-of-the-art model. Specifically, it integrates three core components: an LLM-based Text Grounder, a Candidate Positioning and Setting Marks module, and an LMM-based Visual Grounder. Extensive experiments demonstrate that LLM-Optic outperforms current state-of-the-art methods in a zero-shot setting across multiple datasets. Moreover, to the best of our knowledge, LLM-Optic is the first study to utilize LLMs and LMMs to empower computer vision methods, providing valuable insights for future research.

\clearpage
\small
\bibliography{references}
\bibliographystyle{unsrtnat}

\clearpage
\appendix
\begin{appendices}
   \section{Prompt}
\label{appendix.A}
This section includes the prompts used in the implementation of LLM-Optic and our experiments.
\subsection{Text Grounder}
In the Text Grounder, our goal is to use a Large Language Model (LLM) to analyze user queries and extract the user's underlying intent. The input for this module is the original text query provided by the user, and the output is the refined query. Consequently, we utilize the following prompt, and an example is provided.

\definecolor{darkgreen}{rgb}{0.0, 0.5, 0.0}
\begin{tcolorbox}
    \textbf{Text Grounder Prompt:} \\You are a subject extractor, and you need to extract the subject from the object positioning description I give you. For example, "the painting hanging on the laptop", you need to return to me the real target subject of the sentence "painting". Your answer must be in JSON format. The fixed template is: \{"Subject": "Write answer here, If there are multiple objects, you can use . to divide them. For example chair . person . dog ."\}
    \tcblower
    \small
    \textbf{Query:} \textcolor{teal}{Picture hanging directly above the laptop.}\\
    \textbf{Ground Truth Target:} \textcolor{teal}{Picture.}\\
     \textbf{Text Grounder Output:} \textcolor{teal}{"Subject": Picture.}
\end{tcolorbox}

\subsection{Visual Grounder}
In the Visual Grounder, our goal is to use a Large Multimodal Model (LMM) to determine which marked objects in an image accurately correspond to the target described in the original query. The inputs for this module are the marked image and the original query, while the output is the identifier of the marked object that matches the query. Consequently, we utilize the following prompt, and an example is provided.
\definecolor{darkgreen}{rgb}{0.0, 0.5, 0.0}
\begin{figure}[ht] 
\centering
\begin{tcolorbox}
    \small
    {\centering
    \includegraphics[width=0.5\linewidth]{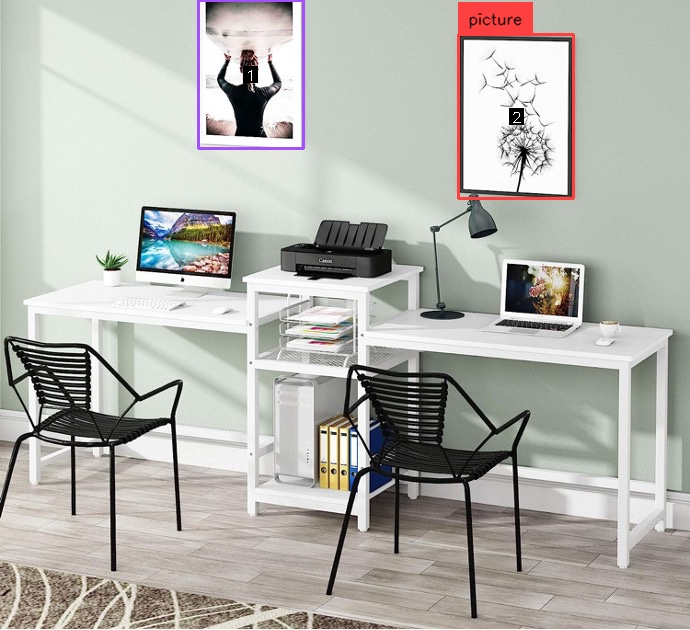}\\}
    \small
    \textbf{Visual Grounder Prompt:} \\You are a target detector. I have marked several candidate targets with boxes and numbers on the input image. I will give you a description and you will choose the target that best fits the description. Your answer must be in JSON format. Answer the number that meets the requirements, that is, the numerical label of the target object. The fixed template is: \{"Subject": "Please provide your answer in the form of an array, for example, [1]. If there are multiple objects, use a comma-separated list within the brackets, such as [1, 3, 4]. If there are no objects in the image that fit the description, return: There are no targets that fit the description."\}
    \tcblower
    \small
    \textbf{Query:} \textcolor{teal}{Picture hanging directly above the laptop.}\\
    \textbf{Ground-Truth Target:} \textcolor{teal}{2}\\
    \textbf{Visual Grounder Output:} \textcolor{teal}{"Subject": [2]}
\end{tcolorbox}
\end{figure}

\subsection{GPT-4V}
In the experiments involving the application of GPT-4V for visual grounding, our goal is to use GPT-4V directly as a visual grounding model. The inputs for GPT-4V are the original input image and original text query, and the outputs are the corresponding bounding boxes of the objects that match the query. Consequently, we utilize the following prompt, and an example is provided.
\begin{tcolorbox}
    \textbf{GPT-4V Prompt:} \\ Please help me find objects that match the description and return the Bounding box in the format of [x, y, w, h]. The format of [x, y, w, h] refers to boxes represented via corner, width, and height, x1, y2 being top left, w, h being width and height. The template for your answer is \{"Subject": "[x, y, w, h]. If you do not answer according to this format, you will be deemed failed."\} 
\end{tcolorbox}

\section{More Experimental Details}
\label{appendix.B}
Due to the \textbf{training-free} nature of LLM-Optic, our framework does not require any GPUs for training, and inference can be performed using either GPUs or CPUs. This significantly lowers the entry barriers to using our framework, enhancing its accessibility. For parameter settings, we employed the default configurations of Grounding DINO and set the temperature to 0.75 for both Text Grounder and Visual Grounder. Across all experiments, the random seed was consistently set to 42 to ensure reproducibility. For the pre-trained models used in the baselines, we utilized UNINEXT-R50 with a ResNet-50 \cite{he2016deep} backbone for UNINEXT and Grounding-DINO-T with a Swin-T \cite{liu2021swin} backbone for Grounding DINO, which are also the default models used for grounding tasks in their respective demos.

In addition to the outstanding performance of LLM-Optic, it is important to note that we employed specific settings to ensure fairness in our experiments. These settings suggest that the actual performance of LLM-Optic may be higher than what is indicated in Table \ref{tab:performance_comparison}. The specific settings are as follows.
\begin{itemize}
    \item Due to the occasional instability of the OpenAI API\footnote{https://openai.com/}, it sometimes returns errors. These errors may be attributable to server network issues or from reaching the request rate limits, etc. To maintain fairness in our experiments, any error returned by the OpenAI API was counted as an incorrect prediction in our LLM-Optic results, resulting in an Intersection over Union (IoU) score of zero. This setting may cause the calculated precision to be lower than the actual value, as these API errors can often be resolved by re-initiating the process.
    
    \item To ensure broad applicability, our prompts were not specifically optimized or fine-tuned for any particular dataset. Customizing prompts to suit different datasets' characteristics could potentially lead to performance improvements.
    
    \item The performance gap between Grounding DINO and LLM-Optic is likely even greater than what the data in Table \ref{tab:performance_comparison} suggest. When confronted with descriptions involving multiple objects, such as "dog not led by ropes outside", Grounding DINO typically detects two bounding boxes corresponding to the dog and the ropes. In the computation of experimental results, we calculate the Intersection over Union (IoU) based on the most probable bounding box, yet the detection of two bounding boxes without semantic understanding is a clear misinterpretation. In contrast, LLM-Optic accurately outputs a single bounding box that correctly corresponds to the intended target, the dog. Although both outcomes are recorded as successes in our experiments, LLM-Optic's outputs are demonstrably more accurate for real-world applications.
\end{itemize}
Alongside these settings, certain inherent limitations of the existing evaluation dataset itself can impact the accuracy of our results. For example, due to ambiguities queries present in the dataset, such as "black shirt, light blue jeans, glasses", our framework may identify multiple objects separately, which are regarded as errors in our evaluation. Furthermore, the RefCOCO series includes descriptions that lack a clear target object, making it difficult for even humans to accurately identify the intended object based solely on these queries. For example, a query like "red top" is ambiguous and poses a significant challenge for our model, thereby constraining its performance. However, such scenarios are rarely encountered in real-world conditions. For specific details and examples, please refer to the Failure Cases (Appendix \ref{appendix.C}). In summary, based on these considerations, the actual accuracy of LLM-Optic is likely higher than what is suggested in Table \ref{tab:performance_comparison}, underscoring its robustness.

\section{Failure Cases}
\label{appendix.C}
\begin{figure*}[t]
	\centering
			{\includegraphics[width=1\linewidth]{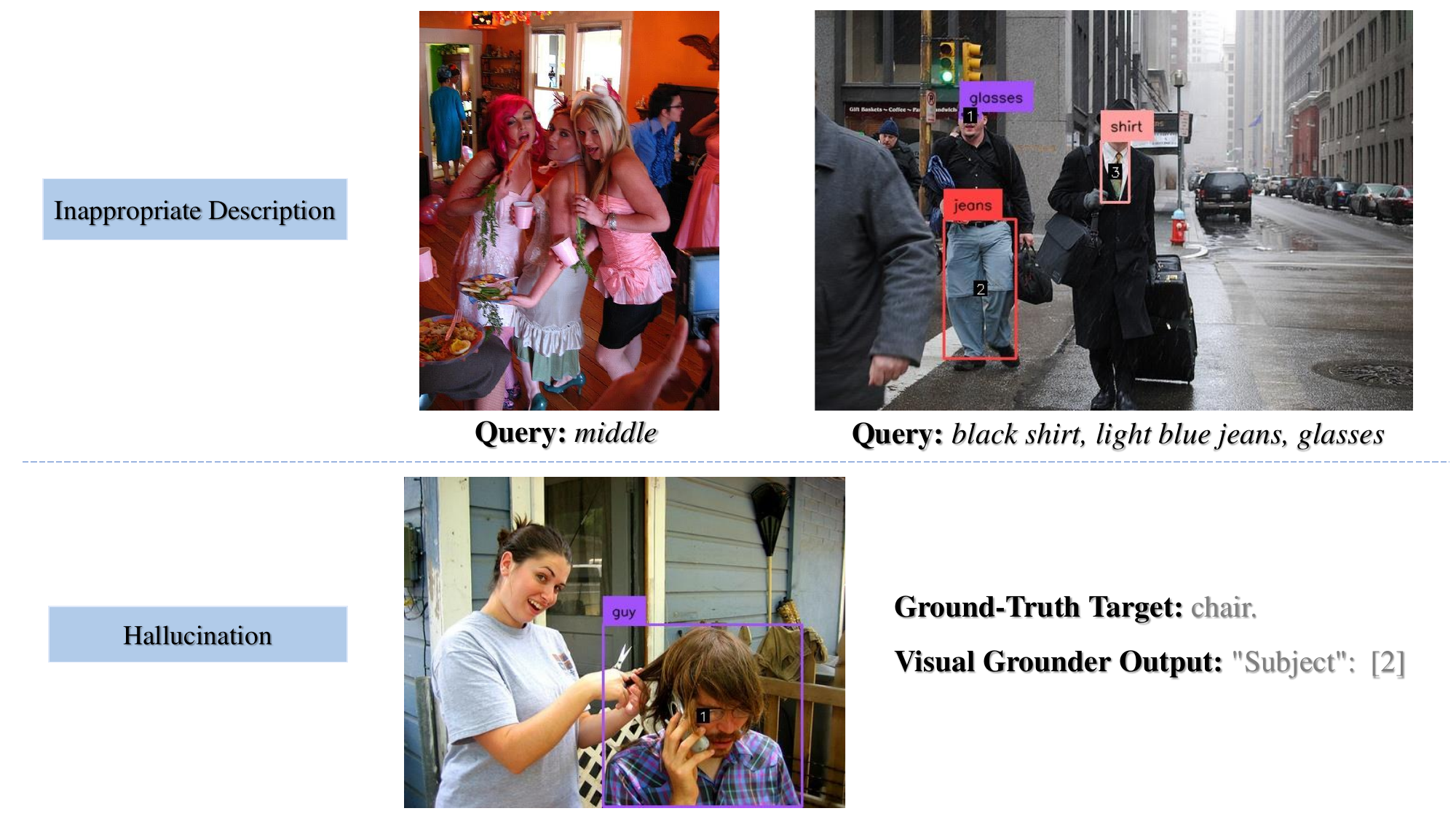}} 
	\caption{\textbf{Failure cases of LLM-Optic.} }
	\label{fig:failurecase}
\end{figure*}

Through the analysis of failure cases, we identified two primary reasons for the prediction failures of LLM-Optic. The first reason is the input of inappropriate descriptions of the target. The second is related to inherent hallucination issues with GPT-4V. Examples of failure cases are presented in Fig. \ref{fig:failurecase}.
\subsection{Inappropriate Description}
When confronted with inappropriate user queries that remain indecipherable even for humans based solely on the query, challenges arise. Specifically, in cases of ambiguous queries, LLM-Optic's Text Grounder is unable to distinguish the target object, potentially resulting in incorrect predictions. For instance, when a query solely consists of descriptions of object appearances and does not provide enough information to discern the intended target, such as "black shirt, light blue jeans, glasses", the Text Grounder is unable to determine the target object from the query alone. Consequently, LLM-Optic identifies three separate targets: a shirt, jeans, and glasses. This leads to failures in detection. Similarly, when a query includes merely a simple location term like "middle" without additional context, it hinders the Text Grounder from identifying the target object, leading to failures.

Facing ambiguous user queries, we discovered that a simple modification to the prompt of the Text Grounder could significantly reduce prediction failures. By merely adding the text "If there is no obvious object, keep the original description unchanged" to the prompt of the Text Grounder, we observed a significant improvement in the results. This finding inspires us to explore more prompt design variations to enhance our performance, which we consider a promising direction for future research.

\subsection{Hallucination}
Hallucination is a widely studied problem within Large Multimodal Models (LMMs). In the module of the Visual Grounder, a very small fraction of errors can be attributed to hallucination. Specifically, GPT-4V sometimes fails to accurately identify markers within an image; for instance, when an object is marked as "1", GPT-4V may incorrectly provide the response "2", even if "1" is the only number present in the image. Due to our modular design, these hallucination issues are expected to be mitigated as LMMs evolve. 

\section{Limitations and Broader Impacts}
\label{appendix.D}
\paragraph{Limitations.}Although LLM-Optic has demonstrated state-of-the-art zero-shot results in visual grounding, it still exhibits certain limitations. Employing LLMs and LMMs for reasoning inevitably increases the OpenAI API cost. This limitation can be mitigated by deploying open-source LLMs and LMMs locally. Moreover, the capabilities of Grounding DINO and LMMs are significant factors constraining our framework's performance. As these techniques improve, the effectiveness of our framework can be further enhanced due to its modular design. Despite these limitations, LLM-Optic establishes a new benchmark in universal visual grounding. To the best of our knowledge, LLM-Optic is the first study to utilize LLMs and LMMs to empower computer vision methods, and it can provide valuable insights for subsequent research.
\paragraph{Broader Impacts.}LLM-Optic is anticipated to have extensive positive societal impacts by significantly enhancing the performance of visual grounding methods in understanding complex queries. This improvement will confer greater robustness to visual grounding, facilitating more effective applications in areas such as autonomous driving and robotics, thereby promoting the advancement of intelligent systems. Consequently, this progress is expected to yield tangible benefits to human life. However, it is also crucial to remain vigilant about the risks of misjudgments and the potential for unemployment associated with deploying automated intelligent systems using this framework.

\section{Additional Results}
\label{appendix.E}
\begin{figure}[htbp]
\centering
\caption*{\textbf{Query}: \textit{a giraffe in between two other giraffes.}}\vspace{-2mm}
\begin{subfigure}{.47\textwidth} 
  \centering
  \includegraphics[width=1\linewidth]{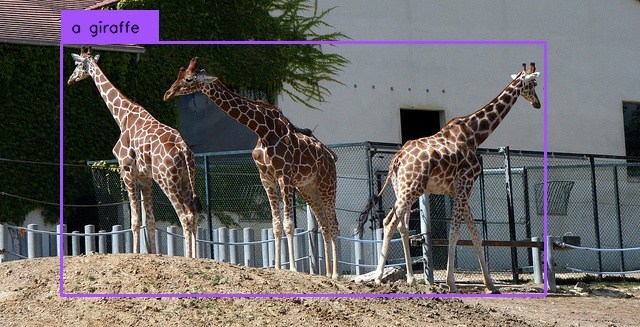}
  \caption*{(a) Grounding DINO}
\end{subfigure}%
\hspace{0.03\textwidth} 
\begin{subfigure}{.47\textwidth}
  \centering
  \includegraphics[width=1\linewidth]{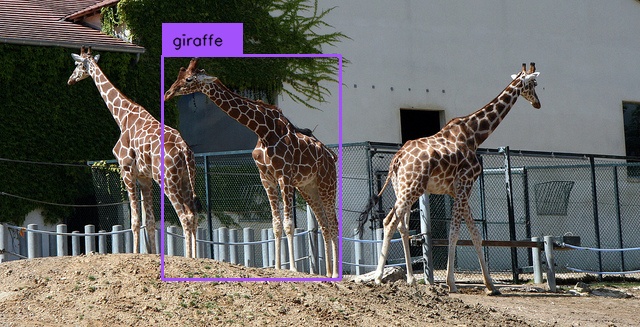}
  \caption*{(b) LLM-Optic}
\end{subfigure}
\vspace{-2mm}

\caption*{\textbf{Query}: \textit{a guy in a red shirt and jeans on top of a brown horse.}}\vspace{-2mm}
\begin{subfigure}{.47\textwidth} 
  \centering
  \includegraphics[width=1\linewidth]{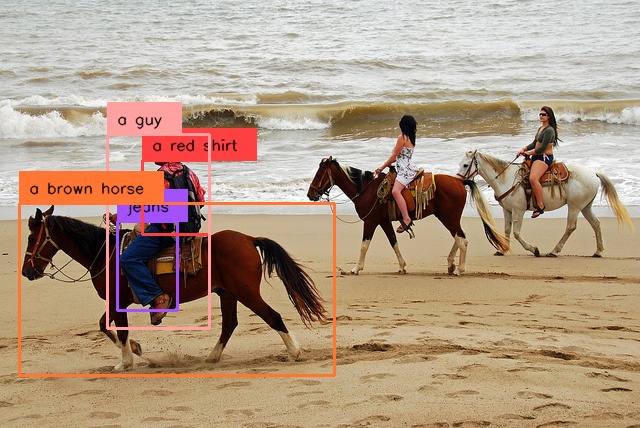}
  \caption*{(a) Grounding DINO}
\end{subfigure}%
\hspace{0.03\textwidth} 
\begin{subfigure}{.47\textwidth}
  \centering
  \includegraphics[width=1\linewidth]{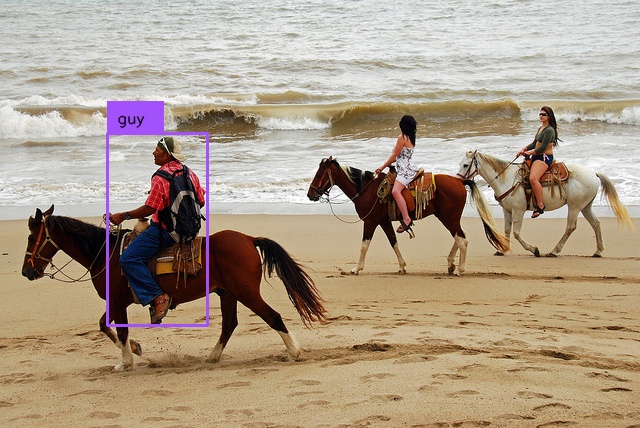}
  \caption*{(b) LLM-Optic}
\end{subfigure}
\vspace{-2mm}

\caption*{\textbf{Query}: \textit{a man wearing red color tshirt pouring wine in glass.}}\vspace{-2mm}
\begin{subfigure}{.47\textwidth} 
  \centering
  \includegraphics[width=1\linewidth]{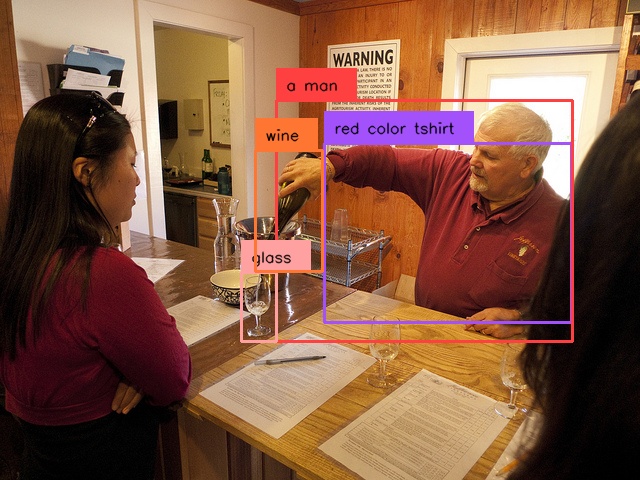}
  \caption*{(a) Grounding DINO}
\end{subfigure}%
\hspace{0.03\textwidth} 
\begin{subfigure}{.47\textwidth}
  \centering
  \includegraphics[width=1\linewidth]{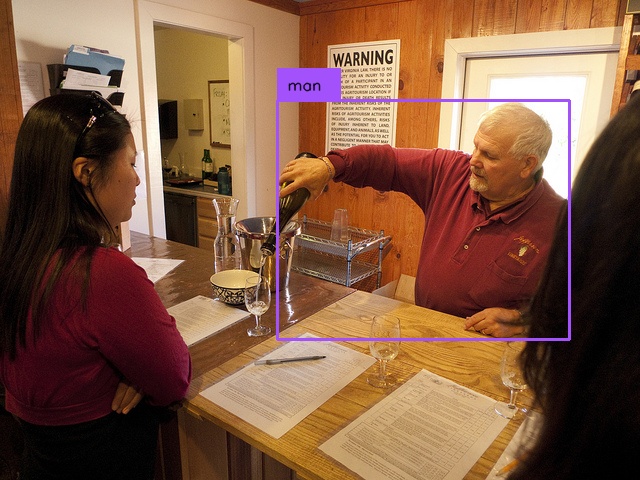}
  \caption*{(b) LLM-Optic}
\end{subfigure}
\vspace{-1mm}
\caption{\textbf{Additional Results (A)}}
\end{figure}

\begin{figure}[ht]
\centering
\caption*{\textbf{Query}: \textit{a woman wearing a sweater and shorts pushing a baby stroller.}}\vspace{-1mm}
\begin{subfigure}{.47\textwidth} 
  \centering
  \includegraphics[width=1\linewidth]{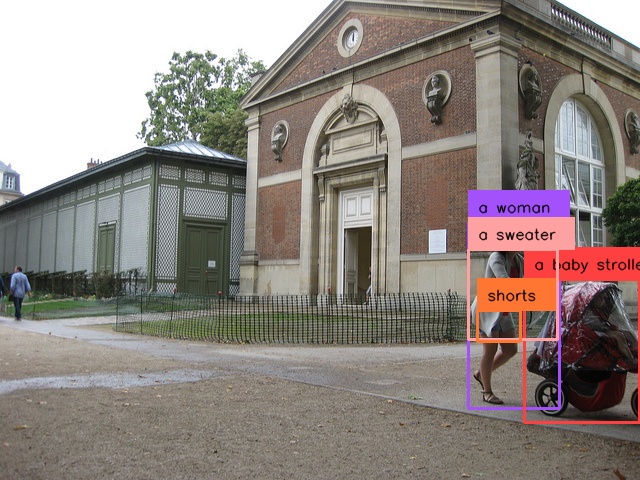}
  \caption*{(a) Grounding DINO}
\end{subfigure}%
\hspace{0.03\textwidth} 
\begin{subfigure}{.47\textwidth}
  \centering
  \includegraphics[width=1\linewidth]{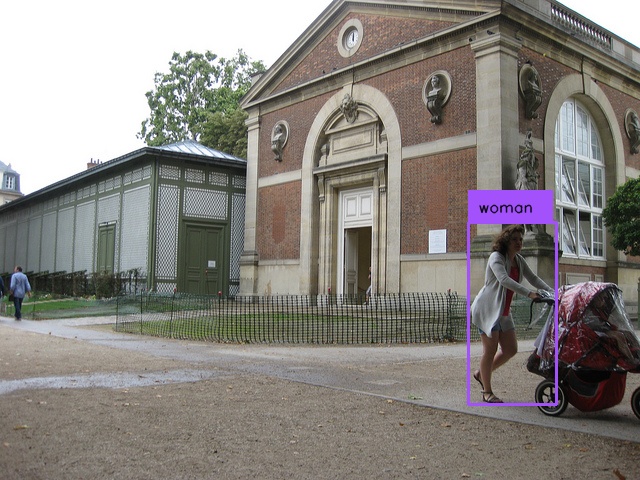}
  \caption*{(b) LLM-Optic}
\end{subfigure}
\vspace{-1mm}

\caption*{\textbf{Query}: \textit{a woman with glasses standing beside a man with glasses.}}\vspace{-1mm}
\begin{subfigure}{.47\textwidth} 
  \centering
  \includegraphics[width=1\linewidth]{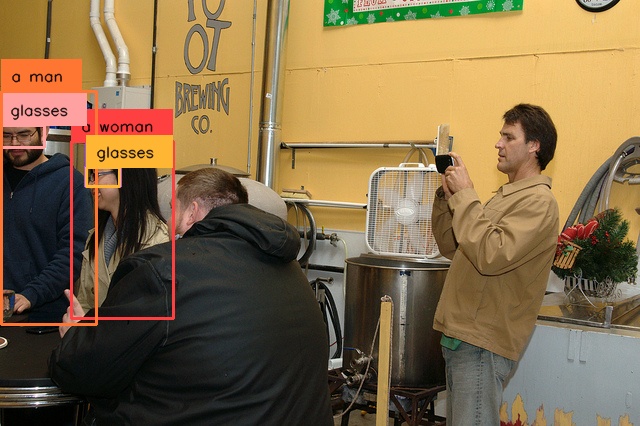}
  \caption*{(a) Grounding DINO}
\end{subfigure}%
\hspace{0.03\textwidth} 
\begin{subfigure}{.47\textwidth}
  \centering
  \includegraphics[width=1\linewidth]{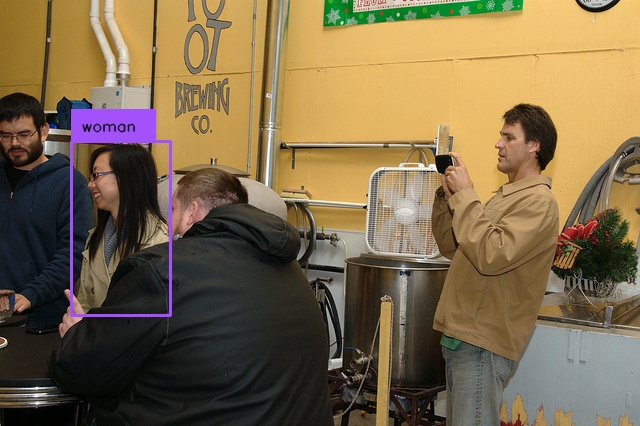}
  \caption*{(b) LLM-Optic}
\end{subfigure}
\vspace{-1mm}

\caption*{\textbf{Query}: \textit{far right jet.}}\vspace{-1mm}
\begin{subfigure}{.47\textwidth} 
  \centering
  \includegraphics[width=1\linewidth]{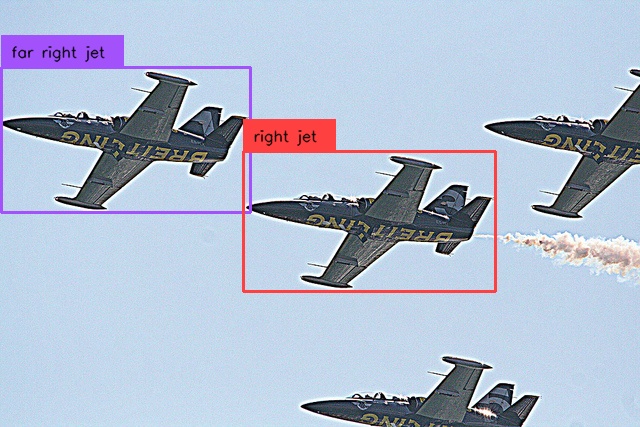}
  \caption*{(a) Grounding DINO}
\end{subfigure}%
\hspace{0.03\textwidth} 
\begin{subfigure}{.47\textwidth}
  \centering
  \includegraphics[width=1\linewidth]{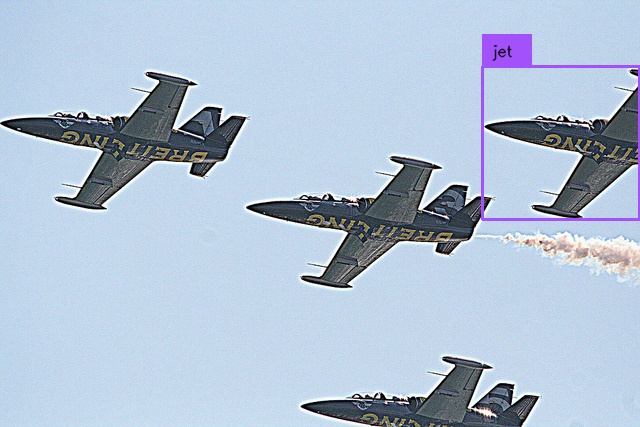}
  \caption*{(b) LLM-Optic}
\end{subfigure}
\vspace{-1mm}

\caption*{\textbf{Query}: \textit{the zebra on the right.}}\vspace{-1mm}
\begin{subfigure}{.47\textwidth} 
  \centering
  \includegraphics[width=1\linewidth]{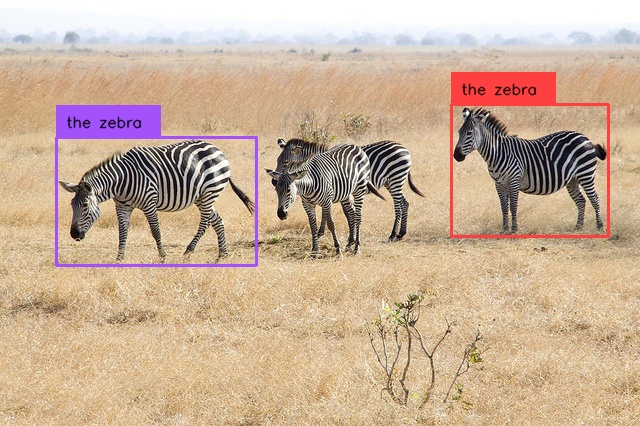}
  \caption*{(a) Grounding DINO}
\end{subfigure}%
\hspace{0.03\textwidth} 
\begin{subfigure}{.47\textwidth}
  \centering
  \includegraphics[width=1\linewidth]{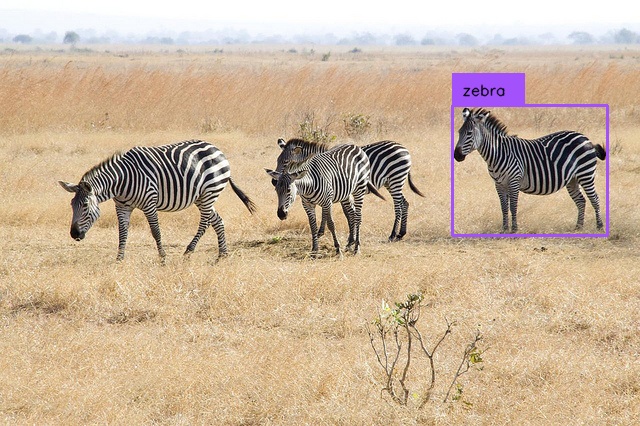}
  \caption*{(b) LLM-Optic}
\end{subfigure} \vspace{-1mm}
\caption{\textbf{Additional Results (B)}}
\end{figure}

\clearpage

\end{appendices}

\end{document}